%% file: acl2021.tex
\pdfoutput=1
\documentclass[11pt]{article}

\usepackage{ACL2021}

\usepackage{times}
\usepackage{latexsym}

\usepackage[T1]{fontenc}
\usepackage[utf8]{inputenc}

\usepackage{microtype}

\usepackage{inconsolata}

\usepackage{placeins}
\long\def\*#1*/{}

\input{utils}

\title{This is not correct! Negation-aware Evaluation of Language Generation Systems}

 \author{Miriam Ansch\"{u}tz \and Diego Miguel Lozano \and Georg Groh \\
         School for Computation, Information and Technology \\
         Technical University of Munich, Germany\\
         \texttt{\{\href{mailto:miriam.anschuetz@tum.de}{miriam.anschuetz}, \href{mailto:diego.miguel@tum.de}{diego.miguel}\}@tum.de}, \texttt{grohg@in.tum.de}
         }
\begin{document}
\maketitle
%\*
\begin{abstract}
%Key message: fine-tuning metrics on our negation dataset yields negation-aware evaluation metric
Large language models underestimate the impact of negations on how much they change the meaning of a sentence. Therefore, learned evaluation metrics based on these models are insensitive to negations. In this paper, we propose \textit{NegBLEURT}, a negation-aware version of the BLEURT evaluation metric. For that, we designed a rule-based sentence negation tool and used it to create the \textit{CANNOT} negation evaluation dataset. Based on this dataset, we fine-tuned a sentence transformer and an evaluation metric to improve their negation sensitivity. Evaluating these models on existing benchmarks shows that our fine-tuned models outperform existing metrics on the negated sentences by far while preserving their base models' performances on other perturbations.
\end{abstract}
\FloatBarrier

\section{Introduction}\label{sec:motivation}
%Semantic faithfulness definition?
%many work has rule-based negation (e.g. for testing negation understanding) but this is not open-sourced or lost in large repos with different objective
%previous work improved negation understanding in NLI and sentiment classification task -> we extend this work to NLG metrics
%Contributions:
%* open-source negation tool
%* large, general-purpose/evaluation focused negation dataset
%* NegMPNet: negation-aware sentence transformer
%* NegBLEURT: negation-aware evaluation metric
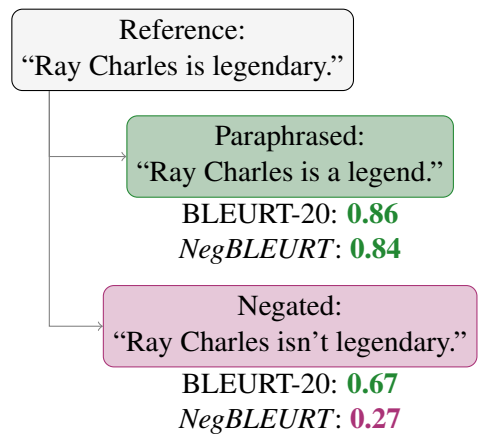
\begin{figure}[t]
    \centering
    \begin{tikzpicture}[align=center]
    \node(orig)[rectangle, draw=black, fill=light, rounded corners] at (0,0) {Reference:\\\enquote{Ray Charles is legendary.}};
    \node(para)[rounded corners, rectangle, draw=darkgreen, fill=lightgreen, below right of =orig, node distance=2cm] {Paraphrased:\\\enquote{Ray Charles is a legend.}};
    \node(bl_o_pos)[below of= para, node distance=.75cm] {\normalsize BLEURT-20: \color{darkgreen} \textbf{0.86}};
    \node(bl_ft_pos)[below of= bl_o_pos, node distance=.5cm] {\normalsize \textit{NegBLEURT}: \color{darkgreen}\textbf{0.84}};
    \node(neg)[rounded corners, rectangle, draw=contrast, fill=lightcontrast, below of =para, node distance=2.25cm] {Negated:\\\enquote{Ray Charles isn't legendary.}};
    \node(bl_o_neg)[below of= neg, node distance=.75cm] {BLEURT-20: \color{darkgreen}\textbf{0.67}};
    \node(bl_ft_neg)[below of= bl_o_neg, node distance=.5cm] {\textit{NegBLEURT}: \color{contrast} \textbf{0.27}};
    \draw[->, draw=gray] ($(orig.south west)+(.5, 0)$) |- (para);
    \draw[->, draw=gray] ($(orig.south west)+(.5, 0)$) |- (neg);
    \end{tikzpicture}
    \caption{Existing metrics like BLEURT-20 fail to score negated sentences correctly. We propose \textit{NegBLEURT} that overcomes this problem while preserving detection performance on other perturbations.}
    \label{fig:vis_abstract}
\end{figure}

Previous work has shown that large language models such as BERT \citep{devlin-bert} lack understanding of negated phrases and do not attribute sufficient importance to the word \enquote{not} \citep{ettinger-bert-negations, hosseini-BERTNOT}. Nevertheless, many widely-used metrics to evaluate natural language generation (NLG) systems, such as BERTScore \citep{zhang-bertscore} or BLEURT \citep{sellam-bleurt}, are based on these models. Automatic evaluation is indispensable when language models are published nearly every day. Moreover, large benchmark datasets make a human evaluation of language models infeasible. Therefore, metric scores are among the most important model selection criteria. However, here we see a severe issue when these metrics fail to distinguish between sentences and their negated versions. Especially when considering fact-checking or entailment prediction, an uninterpreted \enquote{not} can invalidate the entire output of the model and, thus, reduce the trustworthiness of such systems.

While there have been approaches to improve the negation-awareness in natural language inference (NLI) or sentiment classification models \citep{moore-neg-sentiment}, the task of negation-sensitive evaluation of such systems is lacking behind \citep{karpinska-demetr}. An example of such failure is shown in \autoref{fig:vis_abstract}, where a reference sentence is both paraphrased and negated. The well-established BLEURT-20 metric \citep{pu-bleurt20} gives a relatively high score of $0.67$ to this negated sentence, suggesting that it does not fully capture the negation in the sentence.

To extend negation research to the topic of evaluation, we present a negation-aware version of BLEURT, named \textit{NegBLEURT} (\autoref{fig:vis_abstract}). In addition, we released a negation-aware sentence transformer \citep{reimers-sentence-bert} based on an MPNet model \citep{song-mpnet} that extends the application of negation sensitivity to a broader range of tasks. Both models were fine-tuned on a labeled dataset with about $30$ thousand sentence pairs in both their negated and paraphrased versions. We publish this dataset and the sentence negator used to create it together with our models. More specifically, our contributions are:
\begin{itemize}
    \item We open-sourced a rule-based, sentence-level negation tool and released it as a Python package.
    \item Based on this negator, we built a Compilation of ANnotated, Negation-Oriented Text-pairs (CANNOT). This negation evaluation dataset can be used to fine-tune evaluation metrics for negation awareness or probe their sensitivity.
    \item We fine-tuned an MPNet model on our negation dataset. This model returns sentence embeddings that are sensitive to negations.
    \item We published NegBLEURT, a negation-aware version of the BLEURT evaluation metric.
\end{itemize}
Our models were evaluated on various benchmark datasets showing that they greatly outperform their base model on negated sentences while delivering similar performance on other tasks.

\section{Related work}
This section highlights existing work investigating the negation awareness of the BERT language model and different NLG evaluation metrics. Furthermore, we present approaches to improve this awareness with negation pre-training.
\subsection{Studies on negation understanding}
%* General negation in LMs:
%    - \cite{gubelmann-negation-context}: models ARE sensible to negations, findings in other papers are due to missing context
%    - \cite{ettinger-bert-negations}: BERT has terrible insensitivity to negation, completion task w,w/o negation -> BERT prefers affirmation continuation in both cases
%    - \cite{kassner-bert-negations}: BERT bad negation understanding, misprimes; BERT learns negations with supervision but fails in unsupervised setting, negated LAMA dataset
%* Semantic textual similarity assessment
%    - \cite{leung-negation-bertscore}: negation awareness of metrics (bertscore vs. graph-based), metrics unaware but very small test dataset
%* NLG eval metrics:
%    - \cite{karpinska-demetr}: large set of perturbations for MT translations to probe metrics for their sensitivity, negation and antonyms among them
%    - \cite{koch-continuous-perturbation}: eval of common metrics, continuous scale (how does score decrease when perturbations get worse?), negation results bad
%    - \cite{opitz-s3bert}: sentence transformer among multiple dimensions -> (sentences can be close in context but far in negation dimension), explainable scores!
%    - \cite{sai-perturbation}: perturbation checklist for NLG metrics, different seq2seq tasks, antonyms&negation show high deviation from human scores
%    - TODO: add semsim paper
\citet{ettinger-bert-negations} showed that BERT is insensitive to negations. She designed a completion task where the hypernym description of a word was masked. In addition, a \enquote{not} was added to the sentences, resulting in affirmative and negated versions of each sentence. BERT predicted correct hypernyms for both versions, meaning that the model failed to consider the negation indicator. Similar results were achieved by \citet{kassner-bert-negations}. However, they obtained correct completions when the model was fine-tuned on a negation classification task.

To investigate if BERT-based metrics inherit this lack of negation awareness, \citet{leung-negation-bertscore} inspected evaluation metrics such as BERTScore \citep{zhang-bertscore} and Sentence-BERT \citep{reimers-sentence-bert}. They used these metrics to calculate the semantic similarity between 20 equivalent and negated sentence pairs. The BERT-based metrics returned high similarity values, indicating they were robust to negations. More large-scale experiments were performed by \citet{karpinska-demetr} and \citet{sai-perturbation} where different metrics, including BERTScore and BLEURT \citep{sellam-bleurt}, were evaluated on a collection of sentence and word-level perturbations, including negations and antonyms. Both studies show that most suggested evaluation metrics struggle to detect negations and deviate strongly from human evaluations. \citet{koch-continuous-perturbation} examined the robustness towards these perturbations on a continuous scale by gradually introducing more perturbations to the sentences and, hence, decreasing their quality step by step. While the metrics' scores lowered for other perturbations, the scores for the negated sentences remained relatively high, indicating insensitivity towards negation.

\subsection{Improving negation awareness}
%* Datasets:
%    - negated LAMA \cite{hosseini-BERTNOT}
%    - MonotonicityNLi MoNLI \cite{geiger-neg-nli}, negation-aware NLI, FT improves negation performance
%    - NaN-NLI \cite{truong-NaN-NLI}, negated hypothesis, 117 contradicting pairs
%    - WikiFactCheck \cite{sathe-wiki-factcheck}, claim-contex-evidence pairs, manual refuted claims (use negations and antonyms), but go further e.g. with hallucinations
%* Other tasks:
%    - \cite{helwe-TINA}: FT for negated NLi datasets, T5 outperforms on negation without fine-tuning
%    - \cite{moore-neg-sentiment}: investigate multi-task learning (MTL) effects on sentiment classification, "MTL using negation (speculation) as an auxiliary task does make TSA models more robust to negated (speculative) samples"
%* pre-training strategies for negation-aware (BERT) models:
%    - \cite{khandelwal-negbert}: BERT ft for Negation Cue Detection and Scope Resolution
%    - \cite{truong-cue-pretrain}: negation aware masking strategy for better negation cue/span detecion
%    - \cite{hosseini-BERTNOT}: min likelihood of object of negated sentences, BERTNOT as negation-aware BERT
%    - BERT and memorization of negation (https://dke.maastrichtuniversity.nl/jerry.spanakis/wp-content/uploads/2021/11/NegationBERT.pdf)
%-> no approach to improve negation-awareness in evaluation metrics!
Negation awareness is crucial for the task of natural language inference (NLI), in which models predict if two sentences entail or contradict each other. Hence, multiple datasets with negated samples and models trained on them have been published \citep{geiger-neg-nli, helwe-TINA}. As such, \citet{hosseini-BERTNOT} created BERTNOT by training on the negated LAMA dataset with an unlikelihood training objective. Other negation-aware BERT models are NegBERT \citep{khandelwal-negbert} and CueBERT \citep{truong-cue-pretrain}, which were trained for the task of negation cue detection and negation scope resolution. 
Another task that heavily relies on negation awareness is sentiment classification. \citet{moore-neg-sentiment} proposed multi-task learning with a negation speculation auxiliary task to improve the model's performance on negated samples. 

While there has been extensive work on negation understanding in NLI and other tasks, we could not find approaches to improve negation awareness for NLG evaluation metrics. This paper tries to close this gap by pre-training metrics on negated sentences.

\section{Contrastive negation dataset}\label{sec:cannot}
To make an evaluation metric aware of negations, we need a dataset containing pairs of reference and candidate sentences and a label of how well the candidate fits the reference. To have a balanced dataset, we not only need negated, but also meaning-preserving paraphrases of the reference sentence. As described in the previous section, there exist multiple datasets focusing on negations. However, most of these datasets are either targeted towards specific tasks such as NLI or only contain negated sentence pairs. Thus, we processed and aggregated the existent datasets producing a Compilation of ANnotated, Negation-Oriented Text-pairs (CANNOT), which addresses and solves these issues. More specifically, our negation-evaluation dataset is based on the following resources:

\begin{itemize}
    \item \textit{Not another Negation Benchmark} \citep{truong-NaN-NLI}: This dataset was published to improve negation awareness in NLI and includes negated sentence pairs. We filtered for samples with the label \enquote{contradiction}, resulting in $117$ negated pairs.

    \item \textit{Automated Fact-Checking of Claims from Wikipedia} \citep{sathe-wiki-factcheck}: This dataset contains claim-refutation pairs from texts extracted from Wikipedia. The refutation, i.e., the factually incorrect sentence, is often created by negating the claim or replacing one of its words with an antonym. Including more nuanced negations as antonyms and other semantic artifacts diversify the negations in our dataset, making the models trained on it more robust to different negation forms. Nonetheless, many refuted sentences also included further augmentations, such as hallucinations. To only focus on negations, we discarded sentence pairs that had a Jaccard similarity coefficient of less than $0.55$ or differed in length by four or more words. The word splits were obtained with simple white-space tokenization. After the processing, $14,970$ samples were kept.
    
    \item \textit{GLUE Diagnostic Dataset} \citep{wang-glue}: Again, this dataset is targeted to NLI and contains changes beyond pure negation. As with the other datasets, we selected only samples labeled as contradiction and dropped pairs with low Jaccard similarity coefficients or large differences in their lengths. This selection resulted in $154$ samples.
    
    \item \textit{Sentiment-annotated reviews} \citep{kotzias-sentiment-dataset}: This dataset contains online reviews with a strong positive or negative sentiment and, thus, broadens the domains covered by our data. We selected sentences with an auxiliary verb and at most $33$ words. Then, we used our rule-based negation tool (see following \autoref{sec:rule-negation}) to create negated versions of the sentences. In total, $2,110$ further samples were collected.
\end{itemize}

\begin{figure*}
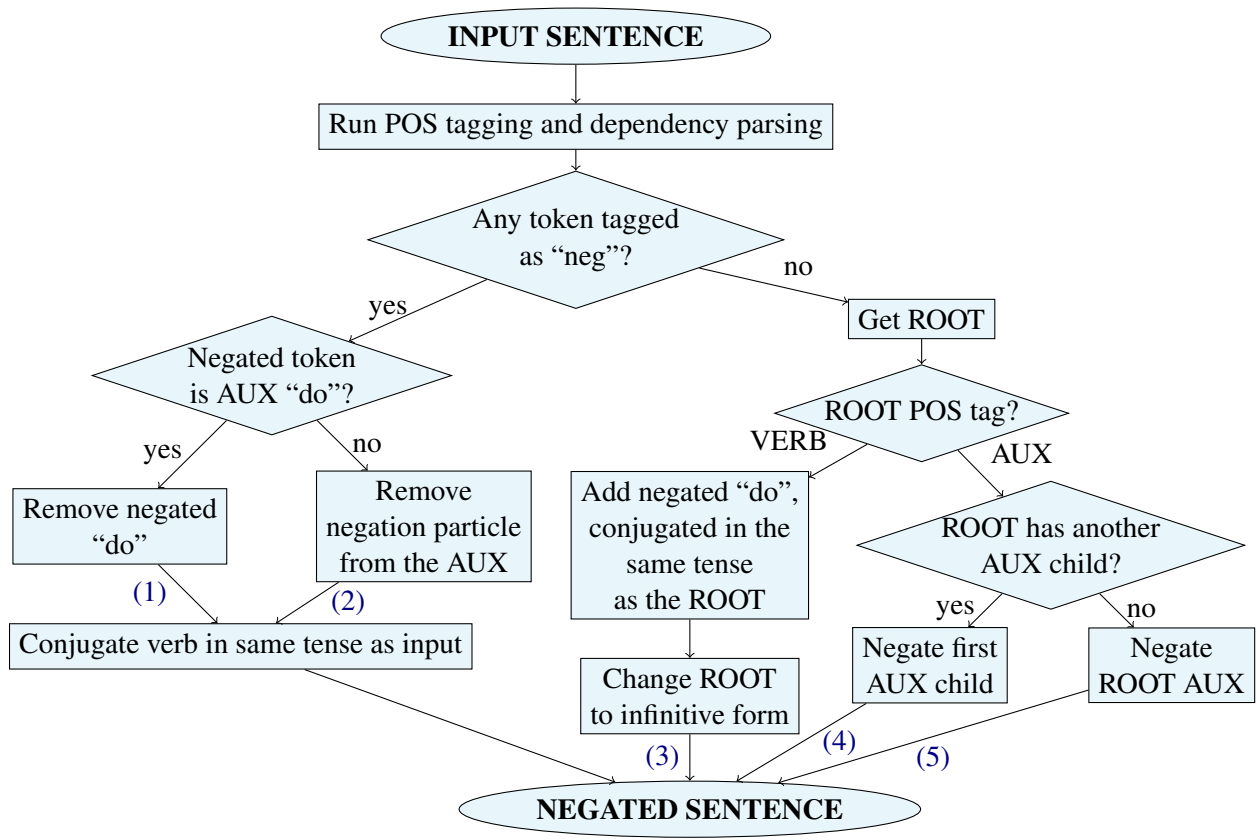

    \centering
    \ruleNegationTikz
    \caption{Flow chart for rule-based sentence negation. The negator can delete negation cues from already negated sentences as well as add them to negate a sentence.}
    \label{fig:rule_neg}
\end{figure*}

\begin{table*}[!ht]
    \centering
    \begin{tabular}{cll}\toprule
        \textbf{Branch} &\textbf{Input sentence} & \textbf{Negated sentence} \\ \midrule
        \color{darkblue}(1) & I \textit{didn't know} what to do. & I \textit{knew} what to do.\\
        \color{darkblue}(2) & I \textit{have never been} to Paris. & I \textit{have been} to Paris. \\
        \color{darkblue}(3) & I \textit{enjoyed} it so much. & I \textit{did not enjoy} it so much. \\
        \color{darkblue}(4) & I \textit{will be} there. & I \textit{won't be} there. \\
        \color{darkblue}(5) & I\textit{'m} very hungry. & I\textit{'m not} very hungry.\\
        \bottomrule
    \end{tabular}
    \caption{Example sentences for different branches in our rule-based negator. Examples (1) and (2) remove a negation from the sentence while examples (3)-(5) add one. The user can decide whether the system should prefer contracted versions like \enquote{won't} instead of \enquote{will not}.}
    \label{tab:neg-examples}
\end{table*}

This resulted in a dataset with negated sentence pairs. To extend it with meaning-preserving paraphrases, we used a PEGASUS model pre-trained for this task \footnote{\url{https://huggingface.co/tuner007/pegasus\_paraphrase}} and created paraphrased versions of each of the references. Finally, the dataset was augmented by adding a swapped version of each pair. This results in a dataset of $68,780$ sentence pairs with equal distribution of negated and equivalent samples.
The pre-processed versions of the underlying datasets and our resulting dataset are publicly available on GitHub \footnote{
    \url{https://github.com/dmlls/cannot-dataset/releases/v1.0}
    %URL removed for double-blind review
}.
\subsection{Rule-based sentence negator}\label{sec:rule-negation}
While previous work used rule-based negation to create negation datasets before, their negators are often not open-source or lost in large repositories with code for the overall goal of the project. Therefore, we publish a lightweight and open-source sentence negation tool as Python module\footnote{\url{https://github.com/dmlls/negate}} that can be used beyond the scope of this paper.

Our negation tool focuses on verbal negations and supports the addition and deletion of negation cues on a sentence level. The flowchart for the negator is shown in \autoref{fig:rule_neg}, accompanied by example sentences in \autoref{tab:neg-examples}. To determine whether a sentence is negated and to distinguish between auxiliary verb forms and common verbs, we first apply the POS tagger provided by the spaCy package \citep{Honnibal-spacy}. A negated sentence is a sentence where a token in the dependency tree is labeled as \enquote{neg} (branches {\color{darkblue}(1)} and {\color{darkblue}(2)} in \autoref{fig:rule_neg}). We differentiate between the auxiliary \enquote{do} and other auxiliary verbs to remove this negation particle. We either entirely remove the negated \enquote{do} (e.g., \textit{don't like} $\rightarrow$ \textit{like}) or remove the negation particle from the auxiliary (e.g., \textit{isn't} / \textit{is not} $\rightarrow$ \textit{is}). Afterward, the remaining verb is conjugated to match the form of the auxiliary\footnote{For verb conjugation, we make use of the module LemmInflect, available at {\urlsize \url{https://github.com/bjascob/LemmInflect}}.}.

To negate an affirmative sentence (branches {\color{darkblue}(3)}-{\color{darkblue}(5)}  in \autoref{fig:rule_neg}), we extract the root verb of the dependency tree. If this verb is a full verb and not an auxiliary, we add a negated \enquote{do} matching the conjugation of the respective verb. If the root is an auxiliary verb, we either negate its first auxiliary child, if any, or otherwise negate the auxiliary itself. The user can decide if the negator should prefer the contracted version, e.g., \enquote{don't}, or write all words separately, e.g., \enquote{do not}.

%\subsection{Generating negations with T5}
%To extend the dataset generated by the rule-based negator with more elaborate and diverse negation, we trained a Flan-T5 \citep{chung-flanT5} model to generate negated versions of the input sentences. The model was fine-tuned on the draft version of the negation dataset that was split into train and validation with a ratio of ???. We utilized cross-entropy loss, an Adam optimizer with a learning rate of $10^{-4}$ and weight decay of $0.01$, and trained for one epoch. In the end, we selected the model with the best ROUGE-L score \citep{lin-rouge} on the validation subset. The resulting model is published on the Hugging Face Hub \footnote{
%    %\url{https://huggingface.co/dmlls/all-mpnet-base-v2-negation}
%    URL removed for double-blind review
%} and produces nuanced and sophisticated negations, e.g., by the use of antonyms.Example outputs are shown in Appendix \ref{sec:appendix-t5}.
%*/ %The T5 was not used for the dataset process -> not described in paper
\subsection{NLG evaluation dataset}
We aimed to make evaluation metrics more sensitive towards negations while preserving their ability to detect other errors. Therefore, we added data from the WMT Metrics Shared Task \citep{bojar-wmt17} to our dataset. This human-annotated data focuses on common errors in machine translation outputs and was used to train multiple evaluation metrics before \citep{sellam-bleurt}. We limited ourselves to the datasets from the years 2015 to 2017 since, upon manual review, the more recent datasets were noisier and contained misannotations. We filtered for samples with a score above $-1$, resulting in $9,264$ labeled samples. Most of the scores range between $0$ and $1$, where a score of $1$ or higher indicates a perfect paraphrase. Our negation data was labeled with a score $0$ for the negated pairs and a score $1$ for the meaning-preserving pairs to match the annotation schema of the WMT data. Finally, we split both the WMT data and our negation dataset into training, evaluation, and test subsets with a ratio of 80:10:10 and combined the respective subsets. This results in training data with $62,435$ samples and test data with $7,804$ samples, with a 12\%-88\% distribution of WMT and negation data both.
\section{Negation-aware models}
We publish two different models fine-tuned on our CANNOT data. On the one hand, we fine-tuned a sentence transformer to return negation-aware sentence embeddings. While the cosine similarity of two embeddings can be applied as an evaluation metric, the embedding representations have broader use cases, e.g., for topic modeling \citep{reimers-sentence-bert}. On the other hand, we fine-tuned BLEURT \citep{sellam-bleurt}, explicitly targeted towards the evaluation task. Both models are published in our GitHub repository\footnote{\url{https://github.com/MiriUll/negation_aware_evaluation}} and on the Hugging Face Model Hub\footnote{NegBLEURT: \url{https://huggingface.co/tum-nlp/NegBLEURT}, NegMPNet: \url{https://huggingface.co/tum-nlp/NegMPNet}}. Our NegBLEURT can also be utilized within the Hugging Face Evaluate library \citep{Wolf-Huggingface}\footnote{\url{https://huggingface.co/spaces/tum-nlp/negbleurt}}. 
\subsection{Sentence Transformer fine-tuning}
Our negation sentence encoder is based on an \texttt{all-mpnet-base-v2}\footnote{\url{https://huggingface.co/sentence-transformers/all-mpnet-base-v2}} model and fine-tuned with the Sentence Transformer library \citep{reimers-sentence-bert}. We trained on our negation training data for one epoch with a batch size of $64$, a learning rate of $2e^{-5}$, and an AdamW optimizer. We utilized a multiple negatives ranking loss to increase the latent distance between correctly paraphrased and negated samples. To create an evaluation metric based on this sentence transformer, embeddings for both the reference and candidate sentence are computed and then scored by their cosine similarity. We call this model the negated MPNet, \textit{NegMPNet}. The cosine-similarity metric based on this model achieves a Spearman correlation of $0.72$ with the ratings in the CANNOT-WMT test set. 
% baseline model score (0.32) here?
\subsection{Negation aware evaluation metric}
BLEURT \citep{sellam-bleurt} is a reference-based NLG evaluation metric that encodes the references and candidates with a BERT model \citep{devlin-bert} and predicts a quality score between $0$ and $1$ with a linear regression layer on top of the BERT model. After pre-training with augmented Wikipedia data, BLEURT was trained on WMT data \citep{bojar-wmt17}.
We chose a BLEURT \citep{sellam-bleurt} metric as the base for our evaluation metric and selected the \texttt{bleurt\_tiny} checkpoint, published as test checkpoint on the official GitHub page\footnote{\url{https://github.com/google-research/bleurt/tree/master/bleurt/test_checkpoint}}. This checkpoint is very lightweight with a hidden size of only $128$, instead of $768$ as in standard BERT models. We used the fine-tuning script provided by the authors, and thus, their original hyperparameters and regression (L2) loss. We fine-tuned on the CANNOT training data for 500 steps, resulting in our final \textit{NegBLEURT} checkpoint. This model has a Spearman correlation of $0.65$ with the scores of our test set. %base model 0.08

\section{Evaluation}
\begin{figure*}[ht]
    \centering
    \includegraphics[width=\textwidth, trim=4.5cm 0cm 0cm 1cm]{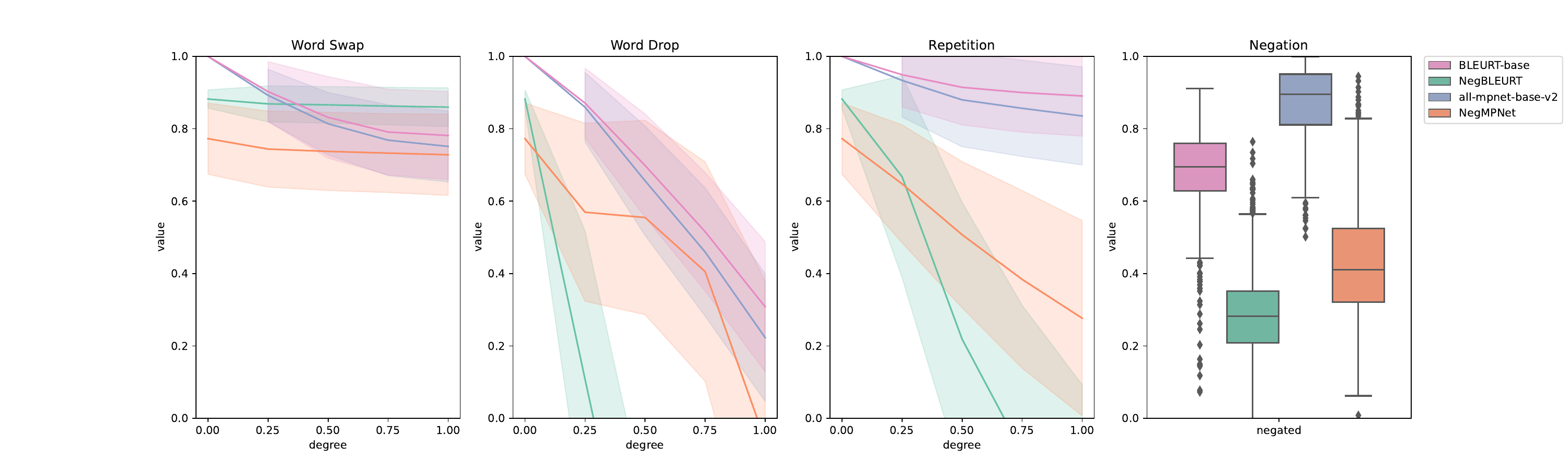}
    \caption{Metrics sensitivities to different degrees of perturbation impairment as introduced by \citet{koch-continuous-perturbation}. Both our proposed models match the performance of their base models on the word swap, word drop and repetition perturbations but clearly outperform them on the negation detection task.}
    \label{fig:metric_comp}
\end{figure*}
A common issue with fine-tuning is catastrophic forgetting \citep{Goodfellow-catastrophic-forgetting}, i.e., the models forget their initial knowledge and overfit the new task. Our fine-tuning approach is successful if it improves negation awareness while retaining performance on other tasks, e.g., not corrupting the detection quality of other errors in candidate sentences. We test our models on common embedding and evaluation benchmarks to test our approach and compare their performances against their respective base models.
\subsection{Massive Text Embedding Benchmark (MTEB)}
MTEB \citep{muennighoff-mteb} is an embedding benchmark that evaluates embeddings on multiple tasks such as classification, clustering, and semantic textual similarity (STS).
It is one of the most extensive collections of tasks, and thus, we evaluated our NegMPNet on this benchmark. As our work targets English, we only evaluate on the English version of the benchmark. The results averaged per task, and the overall macro average are presented in \autoref{tab:mteb_results}, while the performances on the single datasets are provided in \autoref{sec:appendix-mteb}. Unfortunately, some of the datasets (one reranking and seven retrieval) returned errors, and hence, we excluded them from our evaluation. We copied the scores for all-mpnet-base-v2 from the official leaderboard\footnote{\url{https://huggingface.co/spaces/mteb/leaderboard}, as of 17.05.2023}. NegMPNet outperforms its base model in the classification and summarization tasks but shows a decreased performance for clustering, pair classification, and retrieval. When averaging the performances among all tasks, both models perform equally. This benchmark is not targeted towards negation, and, therefore, the results indicate that fine-tuning on our negation data does not harm NegMPNet's general embedding quality.
\begin{table}[ht]
    \centering
    \begin{tabularx}{\linewidth}{Xcc}\toprule
        \multirow{3}{*}{\makecell{\textbf{Benchmark} \\ (num datasets)}} & \multicolumn{2}{c}{\textbf{Model}} \\
        & all-mpnet- & \multirow{2}{*}{NegMPNet}\\
        & base-v2 & \\ \midrule
        \textbf{Average} (60) & 58.78 & 57.16\\ \midrule
        \textbf{Classification} (12) & 65.07 & 70.83\\
        \textbf{Clustering} (11) & 43.69 & 38.45 \\
        \textbf{Pair Classification} (3) & 83.04 & 79.05 \\
        \textbf{Retrieval} (20) & 43.10 & 36.12 \\
        \textbf{Reranking} (3) & 68.83 & 68.24 \\
        \textbf{STS} (10) & 80.28 & 77.58 \\
        \textbf{Summarization} (1) & 27.49 & 29.84\\
        \bottomrule
    \end{tabularx}
    \caption{Comparison of NegMPNet and its base model on the Massive Text Embedding Benchmark (MTEB). We evaluate on different task categories and macro average the scores.}
    \label{tab:mteb_results}
\end{table}
\subsection{Improved negation awareness}
%check underlying datasets -> are there overlaps? no
In this section, we analyze the improved negation awareness beyond the performance on the CANNOT-WMT test set. We selected two NLG evaluation benchmarks that probe negation sensitivity in different metrics and tested NegMPNet (with cosine similarity) and NegBLEURT on them. The results are presented in the following sections.

\subsubsection{Metrics Comparison benchmark}
\citet{koch-continuous-perturbation} probed current, learned metrics for their sensitivity to word swap, repetition, certain word drops, and negations. They gradually increased the level of impairment for all perturbations except negation to measure if the metrics could reflect upon this gradual deterioration. We evaluated our two metrics on their codebase and report the results in \autoref{fig:metric_comp}. NegBLEURT matches the performance of its BLEURT base model \cite{sellam-bleurt} and is sensitive to word drops and repetitions but unaware of word swaps. The same holds for NegMPNet, which shows performances similar to its base model for word swap, word drop, and repetition perturbations. In contrast, NegBLEURT and NegMPNet clearly outperform their base models with a mean difference score of up to $0.5$ for the negation perturbation. These results demonstrate that our models are aware of negation but do not overfit on them and, thus, preserve their performance on tasks aside pure negation detection. 

\subsubsection{DEMETR benchmark}
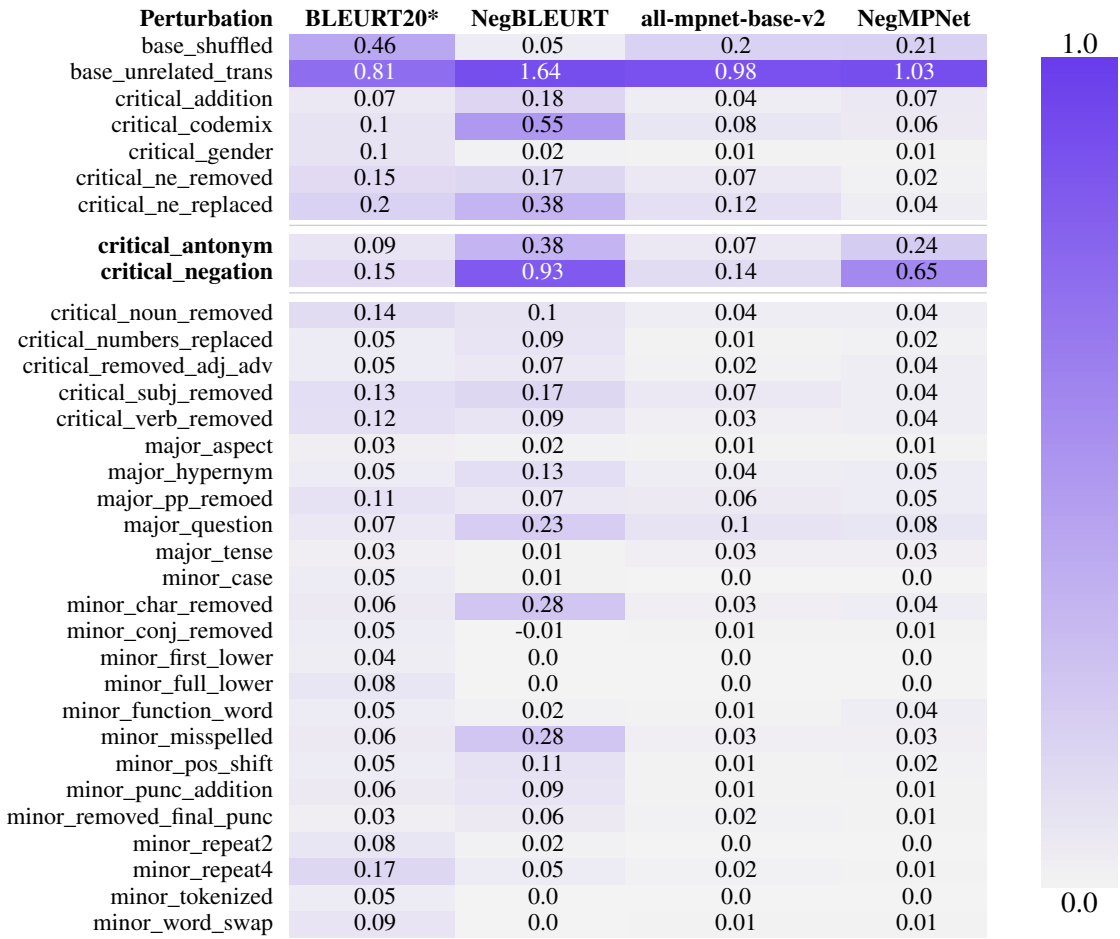
\begin{figure*}[ht]
\begin{minipage}{.85\linewidth}
    \small
    \begin{center}
    \begin{tabular}{r*{4}{c}}
      \textbf{Perturbation}&\textbf{BLEURT20*}& \textbf{NegBLEURT} & \textbf{all-mpnet-base-v2} &\textbf{NegMPNet}\\ 
    base\_shuffled & \gradient{0.46} & \gradient{0.05} & \gradient{0.2} & \gradient{0.21} \\
    base\_unrelated\_trans & \gradient{0.81} & \gradient{1.64} & \gradient{0.98} & \gradient{1.03} \\
    critical\_addition & \gradient{0.07} & \gradient{0.18} & \gradient{0.04} & \gradient{0.07} \\ 
    critical\_codemix & \gradient{0.1} & \gradient{0.55} & \gradient{0.08} & \gradient{0.06} \\
    critical\_gender & \gradient{0.1} & \gradient{0.02} & \gradient{0.01} & \gradient{0.01} \\
    critical\_ne\_removed & \gradient{0.15} & \gradient{0.17} & \gradient{0.07} & \gradient{0.02} \\
    critical\_ne\_replaced & \gradient{0.2} & \gradient{0.38} & \gradient{0.12} & \gradient{0.04} \\ \arrayrulecolor{lightgray}\cmidrule{2-5}
    \textbf{critical\_antonym} & \gradient{0.09} & \gradient{0.38} & \gradient{0.07} & \gradient{0.24} \\ %\cmidrule{2-5}
    \textbf{critical\_negation} & \gradient{0.15} & \gradient{0.93} & \gradient{0.14} & \gradient{0.65} \\ \cmidrule{2-5}
    critical\_noun\_removed & \gradient{0.14} & \gradient{0.1} & \gradient{0.04} & \gradient{0.04} \\
    critical\_numbers\_replaced & \gradient{0.05} & \gradient{0.09} & \gradient{0.01} & \gradient{0.02} \\
    critical\_removed\_adj\_adv & \gradient{0.05} & \gradient{0.07} & \gradient{0.02} & \gradient{0.04} \\
    critical\_subj\_removed & \gradient{0.13} & \gradient{0.17} & \gradient{0.07} & \gradient{0.04} \\
    critical\_verb\_removed & \gradient{0.12} & \gradient{0.09} & \gradient{0.03} & \gradient{0.04} \\
    major\_aspect & \gradient{0.03} & \gradient{0.02} & \gradient{0.01} & \gradient{0.01} \\
    major\_hypernym & \gradient{0.05} & \gradient{0.13} & \gradient{0.04} & \gradient{0.05} \\
    major\_pp\_remoed & \gradient{0.11} & \gradient{0.07} & \gradient{0.06} & \gradient{0.05} \\
    major\_question & \gradient{0.07} & \gradient{0.23} & \gradient{0.1} & \gradient{0.08} \\
    major\_tense & \gradient{0.03} & \gradient{0.01} & \gradient{0.03} & \gradient{0.03} \\
    minor\_case & \gradient{0.05} & \gradient{0.01} & \gradient{0.0} & \gradient{0.0} \\
    minor\_char\_removed & \gradient{0.06} & \gradient{0.28} & \gradient{0.03} & \gradient{0.04} \\
    minor\_conj\_removed & \gradient{0.05} & \gradient{-0.01} & \gradient{0.01} & \gradient{0.01} \\
    minor\_first\_lower & \gradient{0.04} & \gradient{0.0} & \gradient{0.0} & \gradient{0.0} \\
    minor\_full\_lower & \gradient{0.08} & \gradient{0.0} & \gradient{0.0} & \gradient{0.0} \\
    minor\_function\_word & \gradient{0.05} & \gradient{0.02} & \gradient{0.01} & \gradient{0.04} \\
    minor\_misspelled & \gradient{0.06} & \gradient{0.28} & \gradient{0.03} & \gradient{0.03} \\
    minor\_pos\_shift & \gradient{0.05} & \gradient{0.11} & \gradient{0.01} & \gradient{0.02} \\
    minor\_punc\_addition & \gradient{0.06} & \gradient{0.09} & \gradient{0.01} & \gradient{0.01} \\
    minor\_removed\_final\_punc & \gradient{0.03} & \gradient{0.06} & \gradient{0.02} & \gradient{0.01} \\
    minor\_repeat2 & \gradient{0.08} & \gradient{0.02} & \gradient{0.0} & \gradient{0.0} \\
    minor\_repeat4 & \gradient{0.17} & \gradient{0.05} & \gradient{0.02} & \gradient{0.01} \\
    minor\_tokenized & \gradient{0.05} & \gradient{0.0} & \gradient{0.0} & \gradient{0.0} \\
    minor\_word\_swap & \gradient{0.09} & \gradient{0.0} & \gradient{0.01} & \gradient{0.01} \\
    \end{tabular}
    \end{center}    
\end{minipage}
\begin{minipage}{.1\linewidth}
\hspace{8pt}
    \begin{tikzpicture}[scale=1]
    \draw [top color=high!, bottom color = low!, draw=none] (0,0) -- (1,0) -- (1,11) -- (0,11) -- (0,0);
    \node at (.5,11.2) {1.0};
    \node at (.5,-0.2) {0.0};
    \end{tikzpicture}
\end{minipage}
\caption{Sensitivity scores of different NLG metrics on the DEMETR benchmark. The values represent ratios as introduced in \citet{karpinska-demetr}. A higher value denotes a higher sensitivity and is marked in a darker color. Both our models clearly have a higher sensitivity towards negations than their base versions.\newline
\footnotesize{* copied from the original paper}}
\label{fig:demetr_scores}
\end{figure*}
DEMETR \citep{karpinska-demetr} is a diagnosing benchmark dataset for machine-translation output. It contains reference-candidate pairs with different perturbed versions of the candidates, spanning semantic, syntactic, and morphological errors. These errors are categorized by their severity, with categories being critical, major, and minor. The authors measured the sensitivity of the metric to a specific perturbation by predicting the metrics' scores for the reference-candidate and reference-perturbed candidate pairs and calculating the weighted difference between the scores. In their original work, BERTScore achieved the best negation (0.21) and antonym (0.15) detection scores. However, these values still need to catch up to detection scores of other perturbations.

\autoref{fig:demetr_scores} shows the sensitivity scores of our metrics compared to their base versions. Sensitivity scores for other metrics such as BERTScore or COMET are presented on the original DEMETR paper by \citet{karpinska-demetr}. NegBLEURT clearly outperforms all metrics on the critical negation and antonym perturbations while preserving or even improving the detection rate on other perturbations compared to BLEURT20 \citep{pu-bleurt20}. Both all-mpnet-base-v2 and NegMPNet show no sensitivity for most of the perturbations, indicating that these sentence transformers were not trained for the task of NLG evaluation. Nevertheless, NegMPNet shows a competitive detection rate on negations and even antonyms. An increased sensitivity towards antonyms indicates that our fine-tuning approach yields embeddings that distinguish between negated and affirmative sentences beyond the presence of the word \enquote{not}. Although our dataset mainly focuses on simple verbal negations, fine-tuning on it teaches the models to capture antonym-related nuances better. This suggests that our models do not simply learn the artefacts in our CANNOT dataset by hard but can distinguish between different types of contradictions. 

\section{Ablation study}
\begin{table*}[ht]
    \centering
    \begin{tabularx}{\textwidth}{p{3cm}p{1.9cm}XXXp{2.3cm}}\toprule
        \textbf{DEMETR \newline Perturbation} & \textbf{\small Neg-BLEURT} & \textbf{\footnotesize w/o \newline \small Not another Negation Benchmark \citep{truong-NaN-NLI}} & \textbf{\footnotesize w/o \newline \small Automated Fact-Checking of Claims from Wikipedia \citep{sathe-wiki-factcheck}}& \textbf{\footnotesize w/o \newline \small GLUE Diagnostic Dataset \citep{wang-glue}} & \textbf{\footnotesize w/o \newline \small Sentiment-annotated reviews \citep{kotzias-sentiment-dataset} with rule-based negations } \\ \midrule
numbers\_replaced & 0.09 & \gradientrelative{0.06}{0.09} & \gradientrelative{0.04}{0.09} & \gradientrelative{0.11}{0.09} & \gradientrelative{0.05}{0.09}\\ 
gender & 0.02 & \gradientrelative{0.03}{0.02} & \gradientrelative{0.02}{0.02} & \gradientrelative{0.02}{0.02} & \gradientrelative{0.03}{0.02}\\ 
shuffled & 0.05 & \gradientrelative{0.04}{0.05} & \gradientrelative{0.02}{0.05} & \gradientrelative{0.06}{0.05} & \gradientrelative{0.04}{0.05}\\ 
adj\_adv\_removed & 0.07 & \gradientrelative{0.06}{0.07} & \gradientrelative{0.04}{0.07} & \gradientrelative{0.09}{0.07} & \gradientrelative{0.09}{0.07}\\ 
verb\_removed & 0.09 & \gradientrelative{0.1}{0.09} & \gradientrelative{0.06}{0.09} & \gradientrelative{0.01}{0.09} & \gradientrelative{0.11}{0.09}\\ 
noun\_removed & 0.1 & \gradientrelative{0.3}{0.1} & \gradientrelative{0.07}{0.1} & \gradientrelative{0.21}{0.1} & \gradientrelative{0.15}{0.1}\\ 
subj\_removed & 0.17 & \gradientrelative{0.09}{0.17} & \gradientrelative{0.1}{0.17} & \gradientrelative{0.2}{0.17} & \gradientrelative{0.09}{0.17}\\ 
ne\_removed & 0.17 & \gradientrelative{0.14}{0.17} & \gradientrelative{0.1}{0.17} & \gradientrelative{0.17}{0.17} & \gradientrelative{0.15}{0.17}\\ 
codemix & 0.55 & \gradientrelative{0.33}{0.55} & \gradientrelative{0.31}{0.55} & \gradientrelative{0.52}{0.55} & \gradientrelative{0.5}{0.55}\\ 
addition & 0.18 & \gradientrelative{0.2}{0.18} & \gradientrelative{0.05}{0.18} & \gradientrelative{0.18}{0.18} & \gradientrelative{0.17}{0.18}\\ 
antonym & 0.38 & \gradientrelative{0.29}{0.38} & \gradientrelative{0.1}{0.38} & \gradientrelative{0.41}{0.38} & \gradientrelative{0.39}{0.38}\\ 
negation & 0.93 & \gradientrelative{0.74}{0.93} & \gradientrelative{0.35}{0.93} & \gradientrelative{1.02}{0.93} & \gradientrelative{0.82}{0.93}\\ 
ne\_replaced & 0.38 & \gradientrelative{0.26}{0.38} & \gradientrelative{0.25}{0.38} & \gradientrelative{0.4}{0.38} & \gradientrelative{0.4}{0.38}\\ 
        \midrule
        \textbf{Number of \newline removed samples} & 0 & 281 & 53.747 & 399 & 7.475 \\
        \bottomrule
    \end{tabularx}
    \caption{Ablation study of CANNOT subsets. Each data source in the CANNOT-WMT data was removed individually, and the resulting BLEURT checkpoint evaluated on the DEMETR perturbations marked as critical \citep{karpinska-demetr}. Cells with an orange color indicate a decreased performance compared to NegBLEURT, while green indicates an improvement. The darker the color, the larger the difference.}
    \label{tab:ablation_demetr}
\end{table*}
Our CANNOT dataset consists of a diverse collection of datasets, as presented in (\autoref{sec:cannot}). Here, we perform an ablation study to measure the impact of the individual subsets on the model's overall performance. We individually removed each subset from the CANNOT-WMT training and evaluation data, and fine-tuned a new BLEURT checkpoint based on the remaining subsets and the WMT data. The setup for training was the same as for NegBLEURT, which means we fine-tuned the BLEURT-tiny checkpoint for 500 steps using the fine-tuning script from the BLEURT GitHub page. We did not modify the CANNOT-WMT test split, and thus, the number of samples in the subsets deviate from the numbers in \autoref{sec:cannot}.

To compare the impacts of the different datasets, we evaluated the fine-tuned models on all critical perturbations in the DEMETR evaluation benchmark \citep{karpinska-demetr}. The model's sensitivities towards specific perturbations are presented in \autoref{tab:ablation_demetr}. The Wiki-Factcheck dataset \citep{sathe-wiki-factcheck} is by far the biggest subset. As expected, removing it from the training data results in sensitivity drops for nearly all perturbations, especially for antonyms and negations. Removing the \citet{truong-NaN-NLI} dataset from the training data results in a substantial sensitivity loss towards the negation perturbation. This is especially remarkable as the dataset is very small, with less than 300 samples. The BLEURT checkpoint fine-tuned without the GLUE subset shows an increased sensitivity towards negations, indicating that the dataset contains some noise introduced by the selection of contradicting samples. These samples may cover contradictions beyond pure negations that decrease NegBLEURT's performance.

\section{Conclusion}
%negator as simplification module (no negations in Leichte Sprache allowed)
%paragraph level
%different negation positions (not only verb)
%support more languages
In this work,  we created a sentence negation tool that we made available to the research community as a Python package. In addition, we released CANNOT, a data collection for negation detection that can be used to improve negation awareness of language models. We leveraged this dataset to fine-tune a sentence transformer and an NLG evaluation metric. Both models show a strong negation detection ability while preserving task-specific performances compared to their base models. Considering the many papers that pointed out the negation weaknesses of model language models, our work is an important step towards negation awareness.

In the future, we will extend our negator to support more advanced negations beyond the verb level and make our negation dataset multilingual so that multilingual NLG evaluation metrics can be improved as well.

\section{Limitations}
%She's $\rightarrow$ She is or she has?\\
%already/yet etc. in rule-based\\
%only tested on few checklists due to code/data unavailability
%NegMPNet: only few WMT samples means no good metric generalization
Our rule-based negation system works on a verb level and fails for cases that do not match our defined sentence structure. In addition, there are special cases like the sentence \enquote{She's determined} that could use both the verbs \enquote{is} and \enquote{has}. When removing contractions, the negator has to select one of the verbs and may, hence, change the meaning of the input sentence beyond the pure negation. Moreover, in sentences like \enquote{I have not yet been there.}, the adverb \enquote{yet} must be removed or replaced by \enquote{already}, both of which still need to be added to our tool.

We evaluated our negation-aware models on two evaluation metric benchmark datasets. We would have wished to extend this evaluation to further benchmarks, but unfortunately, the lack of published code or datasets, and insufficient code documentation prevented us from doing so.

While NegMPNet achieves remarkable negation detection scores on the considered metric evaluation benchmarks, we must admit that it fails with most of the other perturbations. The all-mpnet-base-v2 model was initially trained as a sentence transformer and not as an evaluation metric, and the small percentage of WMT data in our dataset is insufficient to train it to be such. Therefore, NegMPNet can produce negation-sensitive sentence embeddings but needs further work to be applied as an evaluation metric directly. 

\section*{Ethical Statement}
%negator can help to spread misinformation!
%evaluation important to increase trust in NGL systems
%negator (negation removal) as preprocessing/simplification step
As stated in \autoref{sec:motivation}, trustworthy automatic evaluation metrics are indispensable for selecting and deploying large language models. Metrics that capture negations and reduce the overall score for models that mix up negated and original sentences are, therefore, an important step to increase trust in the metrics themselves, but also in the evaluated models. Moreover, models can be trained to improve negation sensitivity with metrics that detect negation insensitivity, as well as the CANNOT dataset. Therefore, we do not see ethical concerns with our negation-aware metrics or datasets.

However, our negation tool can add or remove a negation to any input sentence. If applied to sentences from the Internet, such as news articles or Twitter posts, it can easily alter the information provided. The negated and original versions still look very similar, and thus, people might oversee the missing or added negation cues when comparing the provided information with other sources. Consequently, we are aware that our negator may be used in a malicious way to spread misinformation. Nevertheless, negation-aware sentence embeddings and evaluation metrics could again detect such modifications. We believe that the benefits of an open-source tool for researchers, as well as the simplified dataset creation it enables, outweigh the drawbacks of potential misuse.

%On the other hand, removing negations from input sentences to machine learning systems can be a useful preprocessing step to improve the system's performance. Moreover, for people with low reading competency, understanding and correctly interpreting negated sentences is often hard. Therefore, the use of affirmative sentences  \citep{plain-negation} %wrong!! our negator is not meaning preserving!

\paragraph*{Supplementary Materials Availability Statement:} All material used in this paper is available to the research community. The sentence negation tool is published as a Python package and in a GitHub repository. The dataset and source code for fine-tuning on this data is also open-sourced on GitHub and Hugging Face. The checkpoints of our models are available on the Hugging Face Model Hub. The links to the individual resources are referenced in their respective paper sections.

\bibliography{custom}
\bibliographystyle{acl_natbib}
%*/
\appendix

\onecolumn
\section{MTEB full results}
\label{sec:appendix-mteb}
\begin{longtable}{clcc}\toprule
        \multirow{2}{*}{\textbf{Task}} & \multirow{2}{*}{\textbf{Benchmark}} &  \multicolumn{2}{c}{\textbf{Model}}\\
        && all-mpnet-base-v2* & NegMPNet\\ \midrule \midrule \endhead
\multirow{12}{*}{\makecell{\textbf{Classification}\\(12 datasets)}}&AmazonCounterfactualClassification & 65.27 & 73.96\\
&AmazonPolarityClassification & 67.13 & 86.1\\
&AmazonReviewsClassification & 31.92 & 41.85\\
&Banking77Classification & 81.86 & 84.23\\
&EmotionClassification & 39.72 & 45.98\\
&ImdbClassification & 70.72 & 68.4\\
&MTOPDomainClassification & 92.08 & 93.38\\
&MTOPIntentClassification & 70.21 & 78.45\\
&MassiveIntentClassification & 69.57 & 74.38\\
&MassiveScenarioClassification & 76.01 & 78.12\\
&ToxicConversationsClassification & 60.86 & 66.15\\
&TweetSentimentExtractionClassification & 55.46 & 58.99\\

\midrule
\multirow{20}{*}{\makecell{\textbf{Retrieval}\\(20 datasets)}}&ArguAna & 46.52 & 19.51\\
&CQADupstackAndroidRetrieval & 56.49 & 53.46\\
&CQADupstackEnglishRetrieval & 52.29 & 49.36\\
&CQADupstackGamingRetrieval & 60.03 & 52.16\\
&CQADupstackGisRetrieval & 44.27 & 40.42\\
&CQADupstackMathematicaRetrieval & 34.21 & 31.8\\
&CQADupstackPhysicsRetrieval & 50.97 & 44.38\\
&CQADupstackProgrammersRetrieval & 44.17 & 41.24\\
&CQADupstackStatsRetrieval & 38.15 & 36.38\\
&CQADupstackTexRetrieval & 33.35 & 30.7\\
&CQADupstackUnixRetrieval & 45.41 & 42.18\\
&CQADupstackWebmastersRetrieval & 44.24 & 42.92\\
&CQADupstackWordpressRetrieval & 35.94 & 33.39\\
&DBPedia & 32.09 & 23.08\\
&FiQA2018 & 49.96 & 26.89\\
&NFCorpus & 33.29 & 27.97\\
&SCIDOCS & 23.76 & 20.1\\
&SciFact & 65.57 & 30.83\\
&TRECCOVID & 51.33 & 58.9\\
&Touche2020 & 19.93 & 16.72\\

\midrule
\multirow{11}{*}{\makecell{\textbf{Clustering}\\(11 datasets)}}&ArxivClusteringP2P & 48.38 & 42.53\\
&ArxivClusteringS2S & 39.72 & 37.92\\
&BiorxivClusteringP2P & 39.62 & 33.7\\
&BiorxivClusteringS2S & 35.02 & 33.45\\
&MedrxivClusteringP2P & 35.58 & 29.97\\
&MedrxivClusteringS2S & 32.87 & 31.48\\
&RedditClustering & 54.82 & 44.31\\
&RedditClusteringP2P & 56.77 & 45.43\\
&StackExchangeClustering & 53.8 & 49.4\\
&StackExchangeClusteringP2P & 34.28 & 30.14\\
&TwentyNewsgroupsClustering & 49.74 & 44.7\\

\midrule
\multirow{3}{*}{\makecell{\textbf{Reranking}\\(3 datasets)}}&AskUbuntuDupQuestions & 65.85 & 65.11\\
&SciDocsRR & 88.65 & 87.75\\
&StackOverflowDupQuestions & 51.98 & 51.87\\

\midrule
\multirow{10}{*}{\makecell{\textbf{STS}\\(10 datasets)}}&BIOSSES & 80.43 & 64.45\\
&SICK-R & 80.59 & 76.71\\
&STS12 & 72.63 & 71.23\\
&STS13 & 83.48 & 84.62\\
&STS14 & 78 & 79.39\\
&STS15 & 85.66 & 84.7\\
&STS16 & 80.03 & 82.17\\
&STS17 & 90.6 & 90.77\\
&STS22 & 67.95 & 57.63\\
&STSBenchmark & 83.42 & 84.18\\

\midrule
\multirow{3}{*}{\makecell{\textbf{PairClassification}\\(3 datasets)}}&SprintDuplicateQuestions & 90.15 & 77.69\\
&TwitterSemEval2015 & 73.85 & 75.98\\
&TwitterURLCorpus & 85.11 & 83.48\\

\midrule
\makecell{\textbf{Summarization}\\(1 datasets)}&SummEval & 27.49 & 29.84\\
\bottomrule
\caption{Detailed performance on MTEB by task and Benchmark dataset. \newline\footnotesize{* copied form the official leaderboard at \url{https://huggingface.co/spaces/mteb/leaderboard}, as of 17.05.2023.}}
\end{longtable}

\end{document}

%% file: utils.tex
\usepackage{csquotes}
\usepackage{hyperref}
\usepackage{tikz}
\usepackage{collcell}
\usepackage{booktabs}
\usepackage{tabularray}
\usepackage{tabularx}
\usepackage{makecell}
\usepackage{longtable}
\usepackage{multirow}
\usepackage[table]{xcolor}
\usepackage{etoolbox}
\usepackage{calc}
\usepackage{pgf} % for calculating the values for gradient
\definecolor{lightblue}{HTML}{e8f6fc}
\definecolor{lightgreen}{HTML}{b6cfba}
\definecolor{darkgreen}{HTML}{228833}
\definecolor{contrast}{HTML}{AA3377}
\definecolor{orange}{HTML}{FF5733}
\definecolor{lightcontrast}{HTML}{e6c8d9}
\definecolor{light}{HTML}{f5f5f5}
\definecolor{white}{HTML}{ffffff}
\definecolor{high}{HTML}{683AEC}  % the color for the highest number in your data set
\definecolor{low}{HTML}{F2F2F2}  % the color for the lowest number in your data set
\newcommand*{\opacity}{90}% here you can change the opacity of the background color!
\newcommand*{\minval}{0.0}% define the minimum value on your data set
\newcommand*{\maxval}{1.0}% define the maximum value in your data set!
\newcommand{\gradient}[1]{
    \ifdim #1pt > \maxval pt
        \cellcolor{high!\opacity} 
    \else
        \ifdim #1pt < \minval pt
            \cellcolor{low!\opacity} 
        \else
            \pgfmathparse{int(round(100*(#1/(\maxval-\minval))-(\minval*(100/(\maxval-\minval)))))}
            \xdef\tempa{\pgfmathresult}
            \cellcolor{high!\tempa!low!\opacity} 
        \fi
    \fi
    \ifdim #1pt > 0.75 pt
        \leavevmode\color{white} #1
    \else
        #1
    \fi
}
\newcommand*{\minvalrelative}{0.0}% define the minimum value on your data set
\newcommand*{\maxvalrelative}{1.0}% define the maximum value in your data set!
\newcommand{\gradientrelative}[2]{ %2 is reference value
    \pgfmathsetmacro{\diff}{#2-#1}
    \ifdim \diff pt > 0 pt
        \pgfmathparse{int(round(100*(\diff/(\maxvalrelative-\minvalrelative))-(\minvalrelative*(100/(\maxvalrelative-\minvalrelative)))))}
        \xdef\tempa{\pgfmathresult}
        \cellcolor{orange!\tempa!white!\opacity} 
    \else
        \pgfmathparse{int(round(100*(-\diff/(\maxval-\minval))-(\minval*(100/(\maxval-\minval)))))}
        \xdef\tempa{\pgfmathresult}
        \cellcolor{darkgreen!\tempa!white!\opacity} 
    \fi
    \ifdim \diff pt > 0.75 pt
        \leavevmode\color{white} #1
    \else
        #1
    \fi
}
\usepackage{tikz}
\usetikzlibrary{shapes.geometric, arrows, positioning, calc}

\newcommand{\ruleNegationTikz}[0]{
    \begin{tikzpicture}[align=center,node distance=2cm] 
    \tikzstyle{myellipse} = [ellipse, draw=black, fill=lightblue]
    \tikzstyle{process} = [rectangle, draw=black, fill=lightblue, align=center]
    \tikzstyle{decision} = [diamond, draw=black, fill=lightblue, aspect=3, align=center, inner sep=1pt]
    \node(inp) [myellipse] {\textbf{INPUT SENTENCE}};
    \node(init)[process] at (0, -1.2) {Run POS tagging and dependency parsing};
    \node(neg_found)[decision, below of = init,node distance=1.5cm] {Any token tagged\\ as \enquote{neg}?};
    
    \node(aux_do)[decision] at (-4,-4.5) {Negated token\\ is AUX \enquote{do}?};
    \node(del_do)[process, below of=aux_do, left of=aux_do] {Remove negated \\\enquote{do}};
    \node(del_part)[process, below of=aux_do, right of=aux_do] {Remove\\ negation particle\\ from the AUX};
    \node(conjugate)[process, below right of =del_do,node distance=2.25cm] {Conjugate verb in same tense as input};
    
    \node(root)[process] at (4.55,-3.75) {Get ROOT};
    \node(root_pos)[decision, below of =root,node distance=1.25cm] {ROOT POS tag?};
    \node(add_do)[process]  at (1.5,-6.75) {Add negated \enquote{do},\\ conjugated in the\\same tense\\ as the ROOT};
    \node(root_inf)[process, below of = add_do] {Change ROOT\\ to infinitive form};
    \node(root_child)[decision, inner sep=0pt] at (6.25,-6.75) {ROOT has another\\AUX child?};
    \node(neg_child)[process, below left of =root_child,node distance=2.25cm] {Negate first\\AUX child};
    \node(neg_root)[process, below right of =root_child,node distance=2.25cm] {Negate\\ ROOT AUX};
    
    \node(out)[myellipse] at (1.5, -10.25) {\textbf{NEGATED SENTENCE}};

    \draw[->] (inp) -- (init);
    \draw[->] (init) -- (neg_found);
    \draw[->]  (neg_found) -- node[anchor=east] {yes} (aux_do);
    
    \draw[->]  (aux_do) -- node[anchor=east] {yes} (del_do);
    \draw[->]  (aux_do) -- node[anchor=west] {no} (del_part);
    \draw[->] (del_do) --node[anchor=east] {\color{darkblue}(1)\hspace{4pt}} (conjugate);
    \draw[->] (del_part) --node[anchor=west] {\color{darkblue}\hspace{5pt}(2)} (conjugate);
    
    \draw[->] (neg_found) -- node[anchor= south west] {no} (root);
    \draw[->] (root) -- (root_pos);
    \draw[->] (root_pos) -- node[anchor=south east] {VERB} (add_do);
    \draw[->] (add_do) -- (root_inf);
    \draw[->] (root_pos) -- node[anchor=south west] {AUX} (root_child);
    \draw[->] (root_child) -- node[anchor=east] {yes} (neg_child);
    \draw[->] (root_child) -- node[anchor=west] {no} (neg_root);
    
    \draw[->] (conjugate) -- (out);
    \draw[->] (root_inf) --node[anchor=east] {\color{darkblue}(3)} (out);
    \draw[->] (neg_child) --node[anchor= west] {\hspace{4pt}\color{darkblue}(4)} (out);
    \draw[->] (neg_root) --node[anchor= north] {\color{darkblue}(5)} (out);
    
\end{tikzpicture}
}

\newcommand\urlsize{\fontsize{8.5pt}{10.2pt}\selectfont}